\documentclass[
    sigconf = true,     
    review = false,      
    screen = true,      
    anonymous = false,   
]
{acmart}

\usepackage{booktabs} 

\setcopyright{none}
\acmConference[GECCO '21]{Genetic and Evolutionary Computation Conference}{July 8--12, 2020}{Lille, France}

\usepackage{amsmath}
\usepackage{xspace}
\usepackage{xcolor}
\usepackage{dsfont}
\usepackage{rotating}
\usepackage{makecell}

\usepackage{amsthm}
\usepackage{amsmath}
\renewcommand{\epsilon}{\varepsilon}

\usepackage[flushleft]{threeparttable}
\usepackage{subfig}
\usepackage{multirow}
\usepackage{amsmath}

\acmDOI{10.1145/3449639.3459406}

\usepackage{comment}

\newcommand{\tabitem}{~~\llap{\textbullet}~~}

\begin{document}
\title[]{The Impact of Hyper-Parameter Tuning for Landscape-Aware Performance Regression and Algorithm Selection}

\author{Anja Jankovic}
\affiliation{
  \institution{Sorbonne Universit\'e, LIP6}
  \city{Paris}
  \country{France}}

\author{Gorjan Popovski}
\orcid{}
\affiliation{%
  \institution{Jo\v{z}ef Stefan Institute}
  \city{Ljubljana} 
 \country{Slovenia}
}

\author{Tome Eftimov}
\orcid{}
\affiliation{%
  \institution{Jo\v{z}ef Stefan Institute}
  \city{Ljubljana} 
 \country{Slovenia}
}

\author{Carola Doerr}
\affiliation{
  \institution{Sorbonne Universit\'e, CNRS, LIP6}
  \city{Paris}
  \country{France}}

\begin{abstract}

Automated algorithm selection and configuration methods that build on exploratory landscape analysis (ELA) are becoming very popular in Evolutionary Computation. However, despite a significantly growing number of applications, the underlying machine learning models are often chosen in an ad-hoc manner. 

We show in this work that three classical regression methods are able to achieve meaningful results for ELA-based algorithm selection. For those three models -- random forests, decision trees, and bagging decision trees -- the quality of the regression models is highly impacted by the chosen hyper-parameters. This has significant effects also on the quality of the algorithm selectors that are built on top of these regressions. 

By comparing a total number of 30 different models, each coupled with 2 complementary regression strategies, we derive guidelines for the tuning of the regression models and provide general recommendations for a more systematic use of classical machine learning models in landscape-aware algorithm selection. We point out that a choice of the machine learning model merits to be carefully undertaken and further investigated.
\end{abstract}


\maketitle

\section{Introduction}
A vast majority of real-world optimization problems are too complex to be defined by an explicit mathematical model, and they require (sometimes very expensive) evaluations to assess and to compare the quality of different alternatives. Classical examples for such problems are typically found in settings where the relationship between decision variables and solution quality cannot usually be established without a computer simulation or a physical experiment, both of which require a significant amount of resources, such as crash tests in automotive industry or clinical trials in medicine. A line of research known as \emph{black-box optimization} (BBO) studies solving those types of problems. They can only be solved using so-called \emph{iterative sampling-based heuristics}, which are algorithms that guide the search towards the optimal solution of a certain problem by sampling candidate solutions, evaluating them (assessing their quality) and using gained information to select the new candidate solutions for the next step. They proceed in iterations until converging to an estimated optimum. 

In the last decades, many iterative optimization heuristics have been designed and improved upon~\cite{Bac96, ES03, EDAMuehlenbein, EDAbook, hansen2001self_adaptation_es, de, KE95}, but the quest for good algorithm designs is far from over. Quite the contrary, the design of efficient iterative sampling-based heuristics is a very active area of research, which conceives considerable attention at the moment due to the important role that black-box optimization techniques play in artificial intelligence. While often being advertised as universal optimization techniques, it cannot be neglected that different algorithms can show very different performances on different types of problems (and for different performance measures). Some -- and often a substantial amount of -- customization is therefore needed to obtain peak performance.

Consequently, a very important and challenging task is to select the most efficient and appropriate algorithm every time one is faced with a new, previously unseen problem instance. This research problem, formalized as the \emph{algorithm selection problem (ASP)}~\cite{RiceASproblem}, has been classically tackled by relying on expert knowledge in both the problem and the algorithm domain. However, significant progress in machine learning (ML) in recent years has allowed for a shift to be made towards \emph{automated} selection (AS)~\cite{KerschkeKBHT18, KerschkeT19, MunozASSurvey, JD20} and configuration (AC)~\cite{BelkhirDSS17}, both being important parts of a more general AutoML research framework~\cite{HutterKV19}, where ML techniques (typically based on supervised learning approaches, such as regression and classification) are used to design and train models to predict the performance of different black-box algorithms on unseen problem instances as accurately as possible. These performance predictions would then allow for selecting and/or configuring the best algorithm for the problem at hand. 

One of the challenges for such supervised learning methods to be useful is the identification of convenient representations of the problem instances, which can be used by the AutoML techniques to derive a predictive model. Such representations are usually expressed as vectors of numerical values, each quantifying a relevant characteristic of the instance through an appropriate measure. These numerical representations serve to identify different problem instances and allow to discriminate between them. In evolutionary computation (EC) terms, these quantified measures, known as \emph{features}, thus describe the \emph{fitness landscape} of a problem instance.

Fitness landscapes have been and still are extensively studied in EC. For a long time, the only accessible landscape properties were high-level and intuitive (e.g., degree of multimodality, separability, number of plateaus), and their major drawback is that they all require prior expert knowledge and thus cannot be automatically computed. More recently, however, a line of research under the umbrella term of \emph{exploratory landscape analysis (ELA)}~\cite{mersmann_exploratory_2011} has come up with new ways of proper numerical feature extraction (or more precisely approximation, as we have at our disposal only the samples that we have evaluated and to which a solution quality has been assigned). In addition to the original features sets proposed in~\cite{mersmann_exploratory_2011}, many new features sets, each quantifying some relevant problem characteristic, have since been incorporated~\cite{kerschke2015nbc, lunacek_dispersion_2006, munoz_exploratory_2015, Derbel19FOGA}.  These insights have fast given rise to nowadays very active research questions around the \emph{landscape-aware} algorithm selection and/or configuration.

Since the EC community itself is traditionally predominantly focused on the development or the assessment of different landscape features and search heuristics, the relevance of selecting an appropriate machine learning model needed to provide information for the algorithm selection (AS) is often neglected.  With the exception of selected communities such as AutoML, most authors, following common practices, typically apply easily accessible, off-the-shelf techniques such as default implementations of random forests (RF), support vector machines (SVM), or decision trees (DT) -- available in the \texttt{scikit} library~\cite{pedregosa2011scikit}, for example.  As suggested in~\cite{HutterXHL14}, random forests are considered to be the preferred technique when it comes to landscape-aware algorithm performance prediction, as it was empirically demonstrated that they outperformed other frequently used ML models (e.g., ridge regression, abovementioned SVMs, Gaussian processes, neural networks) in the context of combinatorial and mixed-integer problems. However, typically only a single ML model is trained to perform the task.

With that being said, hyper-parameter tuning of regression models might have a potential of largely improving the regression quality.  In this work, we thus raise the question of the magnitude of impact that hyper-parameter tuning can have on both regression quality and algorithm selection, as the effects of different ML frameworks and their respective strengths and weaknesses have yet to be investigated in detail in the scenario of landscape-aware AS.  We aim to highlight said potential of hyper-parameter tuning on a selected set of tree-based regression models, while arguing that the model's performance quality is in fact highly dependent on the problem set and the algorithm portfolio, which are case-specific.

\paragraph{\textbf{Our Contribution and Results}} 
We compare in this work the performance of 30 different regression models on the algorithm selection task suggested in~\cite{KerschkeT19}. After a preliminary step of testing the quality of the following seven families of regression models with different hyper-parameter values tested via iterative grid search: Random Forests~\cite{breiman2001random}, Decision Trees~\cite{breiman1984classification}, Bagging decision trees~\cite{breiman1996bagging}, Lasso~\cite{tibshirani1996lasso}, ElasticNet~\cite{zou2005regularization}, KernelRidge~\cite{murphy2012machine}, and PassiveAggressive~\cite{crammer2006online}, we retained only the models that largely outperformed the rest in terms of regression quality.  The 30 selected models are all \emph{tree-based}, and they are configurations of random forest, decision tree and bagging decision tree regression techniques.  Note that, due to the rather small size of our data set, we have not considered techniques such as Neural Networks which are powerful for a huge quantity of data.  We have not considered any classification techniques either (including Bayes classifier), as we are interested in predicting numerical performance values for each algorithm.

We train the 30 models on different data sets, taking as input landscape (ELA) features of a problem and outputting the fixed-budget performance of an algorithm.  Two distinct ELA feature representations (based on different sample sizes for feature computation) describe each of 120 problem instances belonging to 24 problem classes that were used in our paper. Fixed-budget performances for a portfolio of 12 algorithms were recorded for different budgets of function evaluations (we consider budgets of 250, 500, and 1\,000 function evaluations), measuring the target precision (i.e., the distance to the optimum).  On top of that, two complementary regression approaches were adopted for each model and each data set: one that predicts true (unscaled) target precision values and another that predicts $\log_{10}$ target precision, as suggested in~\cite{JankovicD20}.

We then use stratified 5-fold cross-validation to ensure that all problem instances were used in the test phase of our models, and build for each regression model and each data set an automated algorithm selector, which takes as input the predicted target precision of each of the twelve algorithms and which returns the algorithm with the best predicted performance. The true target precision of this selected algorithm is then compared to that of the actual best algorithm for that problem instance, which defines the loss that we associate to the regression model.
Following the approach in~\cite{JankovicD20}, for each regression model and each data set, we also build an algorithm selector which combines the regression for the unscaled target precision with that for the logarithmic precision, favoring the latter for small target precision values, and favoring the former otherwise. 
In total, we evaluate for each of the 30 regression models, three different algorithm selectors (unscaled, logarithmic, combined), on 6 different data sets (3 different budgets of function evaluations and 2 different feature sample sizes).

The results of our study clearly indicate a need for appropriate tuning of the regression techniques, but more importantly, they raise an important question of how the chosen ML model can lead to highly varying results in the algorithm selection step.  We argue that the untapped potential of a careful choice of the relevant ML model and its hyper-parameter configuration can be only fully exploited when taking into account some preliminary knowledge about the problem classes and the algorithms. This sub-explored area of research merits further investigation.
We see differences of up to several orders of magnitude in the Root Mean Squared Error values of the different models, not aggregated across optimization algorithms. Even if we aggregate the errors, the differences between them are still as high as 60\%.  We further notice that different models perform differently on different types of problems, making it difficult to derive a general recommendation for which model to favor in which scenario. This, however, does not limit the relevance of our work, since training the different regression models is of negligible cost, in particular when compared to the efforts required for setting up the whole algorithm selection pipeline.  In practice, the use of several ML techniques at the same time (``ensembles'') is not uncommon -- quite the contrary, in fact~\cite{EnsembleRegression}.
We therefore also hope that our work motivates further investigations of ensemble learning techniques in the context of landscape-aware algorithm selection and configuration.

\paragraph{\textbf{Related Work}}

Existing research in automated algorithm selection and configuration can be roughly positioned on one of the two main axes, depending on whether machine learning techniques utilized are supervised or unsupervised. In terms of unsupervised learning, reinforcement learning is the most predominant (see~\cite{BiedenkappBHL19, KerschkeHNT19} and references mentioned therein). When it comes to the the supervised approaches, approaches building upon exploratory landscape analysis (ELA)~\cite{mersmann_exploratory_2011} are most frequent in the field. In particular, ELA-based regression have been applied to assess the effect problem features have on algorithm performance~\cite{LiefoogheDVDAT20}, to configure algorithms' parameters (per-instance algorithm configuration, PIAC)~\cite{BelkhirDSS17}, as well as to select algorithms from a given portfolio~\cite{KerschkeT19, MunozKH12prediction}. See~\cite{kerschke2018survey, MunozASSurvey} for comprehensive surveys of automated AS state-of-the-art methods and results. With regards to dynamic algorithm selection and configuration, search behavior or algorithms’ state parameters can also influence the model recommendations, see~\cite{BajerPRH19, DerbelLVAT19} and references mentioned therein. 

\section{Performance Regression}
\label{sec:ml-regression}

\subsection{Experimental Setup}
\label{ssec:exp-setup}

\paragraph{Regression Models}
Both regression and classification as supervised learning techniques have been studied in the context of landscape-aware algorithm selection. The advantage of regression models over classification ones is in keeping track of the magnitude of differences between performances of different algorithms, as they predict numerical performance values.

Following best practices in this line of work, as mentioned above, for our analysis we have selected three different classes of regression models, namely Decision Tree~\cite{breiman1984classification}, Random Forest~\cite{breiman2001random} and Bagging Decision Tree~\cite{breiman1996bagging}. For each model class, the hyper-parameter configurations used are shown in Table~\ref{tab:hyperparameters_regression}.

Since the basic component unit of all considered models is a decision tree (both Random Forests and Bagging DTs are based on them), the hyper-parameter \emph{crit} value can be one of the following three: \emph{mse} (mean squared error), \emph{mae} (mean absolute error) and so-called \emph{friedman mse} - Friedman mean squared error. The \emph{minsplit} hyper-parameter represents the minimum number of data instances a tree node has to contain in order to become a splitting node. Lastly, the \emph{nest} hyper-parameter defines the number of decision trees needed to build a Random Forest or Bagging DT model. In total, we end up with 30 different regression models, with 6 different configurations for Decision Tree, 12 for Random Forest, and 12 for Bagging DT. Table~\ref{tab:hyperparameters_regression} summarizes the chosen hyper-parameter values for different regression classes.

\begin{table}[ht]
    \caption{Hyper-parameter values for the regression models}
    \label{tab:hyperparameters_regression}
    \centering
    \resizebox{.4\textwidth}{!}{ 
    \begin{tabular}{l|l}
    \hline
    Model & Hyper-parameters\\
    \hline
        DecisionTree & \tabitem  $crit \in \{"mse", "mae", "friedman\_mse"\}$ \\
         (6 configs.) & \tabitem  $minsplit \in \{4,5\}$ \\
         \hline
        RandomForest & \tabitem  $crit \in \{"mse", "mae"\}$ \\
         (12 configs.)& \tabitem  $minsplit \in \{4,5\}$ \\
         & \tabitem $nest \in \{3,6,9\}$ \\
        \hline
        BaggingDT & \tabitem  $crit \in \{"mse", "mae"\}$ \\
         (12 configs.)& \tabitem  $minsplit \in \{4,5\}$ \\
         & \tabitem $nest \in \{3,6,9\}$ \\

        \hline
    \end{tabular}
   }
\end{table}

\begin{figure*}[t]
    \centering
    \includegraphics[width=\linewidth]{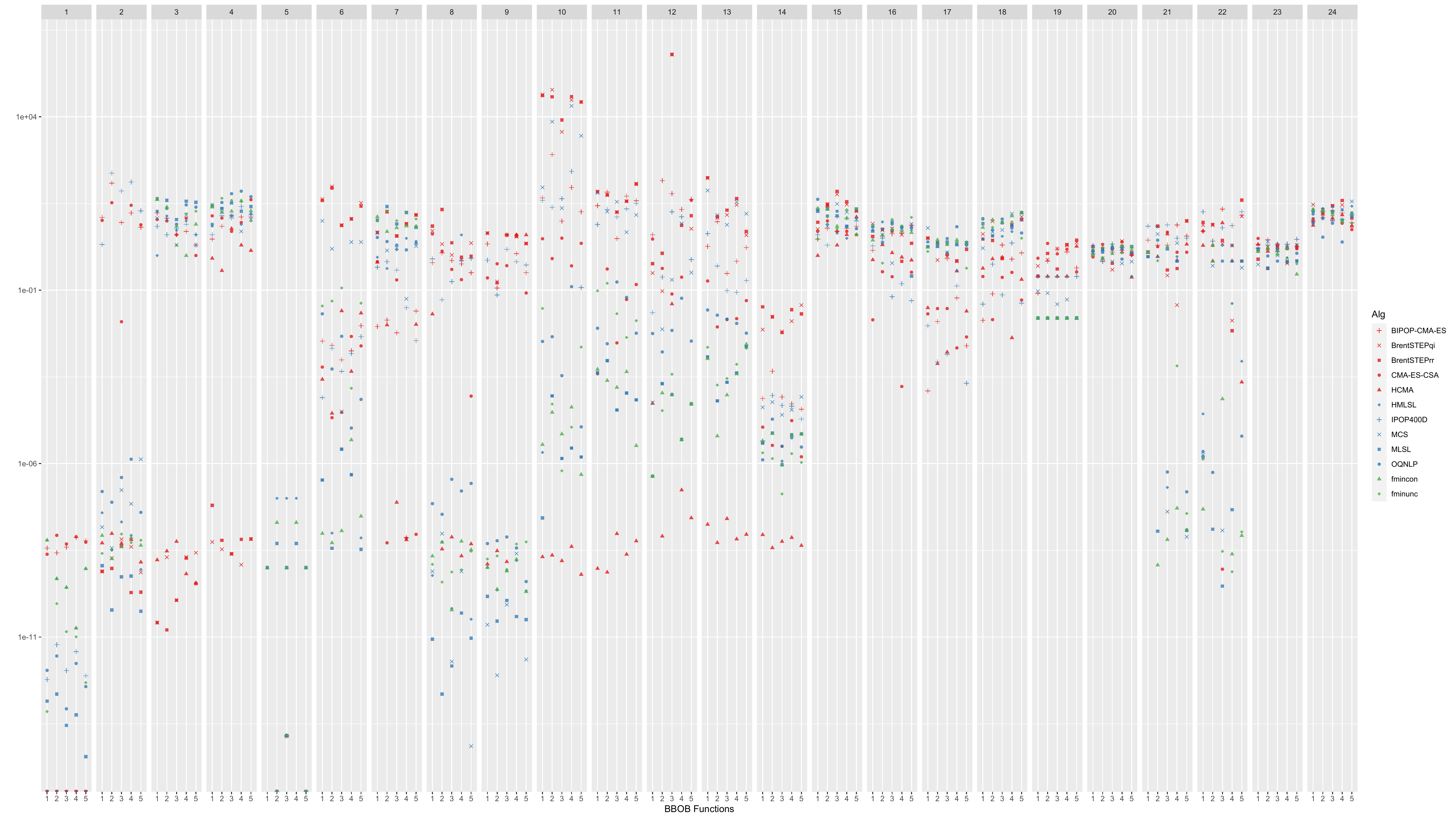}
    \caption{Single-run target precision of the 12 algorithms on the first five instances of 24 BBOB functions. They represent our portfolio from which we want to select the best-performing algorithm for an unseen problem instance.}
    \label{fig:actual-performance}
\end{figure*}

\paragraph{Benchmark problems}
As commonly used in the continuous optimization community, our benchmark set of choice when measuring different algorithm performances was the noiseless BBOB testbed~\cite{bbob-functions} on the COCO platform~\cite{cocoplat}, which is a dedicated environment for comparison of algorithm performance in continuous black-box optimization. It contains 24 different single-objective functions (noted here as FID 1-24), and different \emph{instances} can further be generated based on those 24 initial definitions by rotation and translation in the objective space. We hence consider the first 5 instances of each function (IID 1-5), for a total of 120 problem instances. Lastly, we restrict our analysis to dimension 5.

\paragraph{Algorithm Portfolio}
The algorithm portfolio we chose for this work was suggested in~\cite{KerschkeT19} for its diversity. It consists of the following 12 algorithms: BrentSTEPqi~\cite{PosikB15}, BrentSTEPrr~\cite{PosikB15}, CMA-ES-CSA~\cite{Atamna15}, HCMA~\cite{loshchilov_intensive_2013}, HMLSL~\cite{pal2013benchmarking}, IPOP400D~\cite{auger2013benchmarking}, MCS~\cite{huyer2009benchmarking}, MLSL~\cite{pal2013benchmarking}, OQNLP~\cite{pal2013comparison}, fmincon~\cite{pal2013comparison}, fminunc~\cite{pal2013comparison}, and BIPOP-CMA-ES~\cite{hansen_benchmarking_2009}. Note that, due to the unavailability of the raw performance data for one of the algorithms in the original study, the BIPOP-CMA-ES was added instead of the missing one. The performance data of all twelve algorithms can be downloaded at~\cite{BBOBdata}, but for our setting it was more convenient to extract the relevant figures from IOHprofiler~\cite{IOHprofiler}.  Throughout this work, we focus on the fixed-budget performance, which is a scenario most often seen in real-world application where we can allow ourselves only a limited budget of function evaluations, and where the performance metric used is \emph{target precision} of an algorithm, i.e., the distance between the best solution reached after a certain budget of function evaluations and the estimated optimal solution. Concretely, we consider 3 different budget sizes of 250, 500 and 1000 function evaluations across all algorithms from the portfolio for purpose of sensitivity analysis, and we restrict ourselves to a single algorithm run per problem instance. We show in Figure~\ref{fig:actual-performance} the portfolio's target precisions reached after 1000 evaluations. We note that the algorithm performances are significantly less diverse for functions 15 through 20, 23 and 24 than for the other problems.

\paragraph{Problem features}
Predictor variables for our regression models are vectors of ELA feature values, which quantify the landscape properties of each problem instance. Feature computation is done using the \emph{flacco} package~\cite{flacco} for two distinct sets of uniformly sampled points and their evaluations. Samples are of sizes $50d$ (250) and $400d$ (2000) respectively for purpose of sensitivity analysis. 50 independent feature computations were performed for each sample size, as some of them can show low robustness on certain features~\cite{Renau2019features}, and a median feature value was taken for each one.  Following suggestions from~\cite{KerschkeT19, BelkhirDSS17}, we choose only those feature sets that do not require additional sampling during their computation; this way, we end up with a total of 56 feature values per problem instance, belonging to the following sets: classical ELA (y-Distribution, Levelset and Meta-model)~\cite{mersmann_exploratory_2011}, Dispersion~\cite{lunacek_dispersion_2006}, Information Content~\cite{munoz_exploratory_2015} and Nearest-Better Clustering~\cite{kerschke_detecting_2015} feature sets.

\begin{figure*}[ht!]
    \centering
    \includegraphics[width=0.95\linewidth]{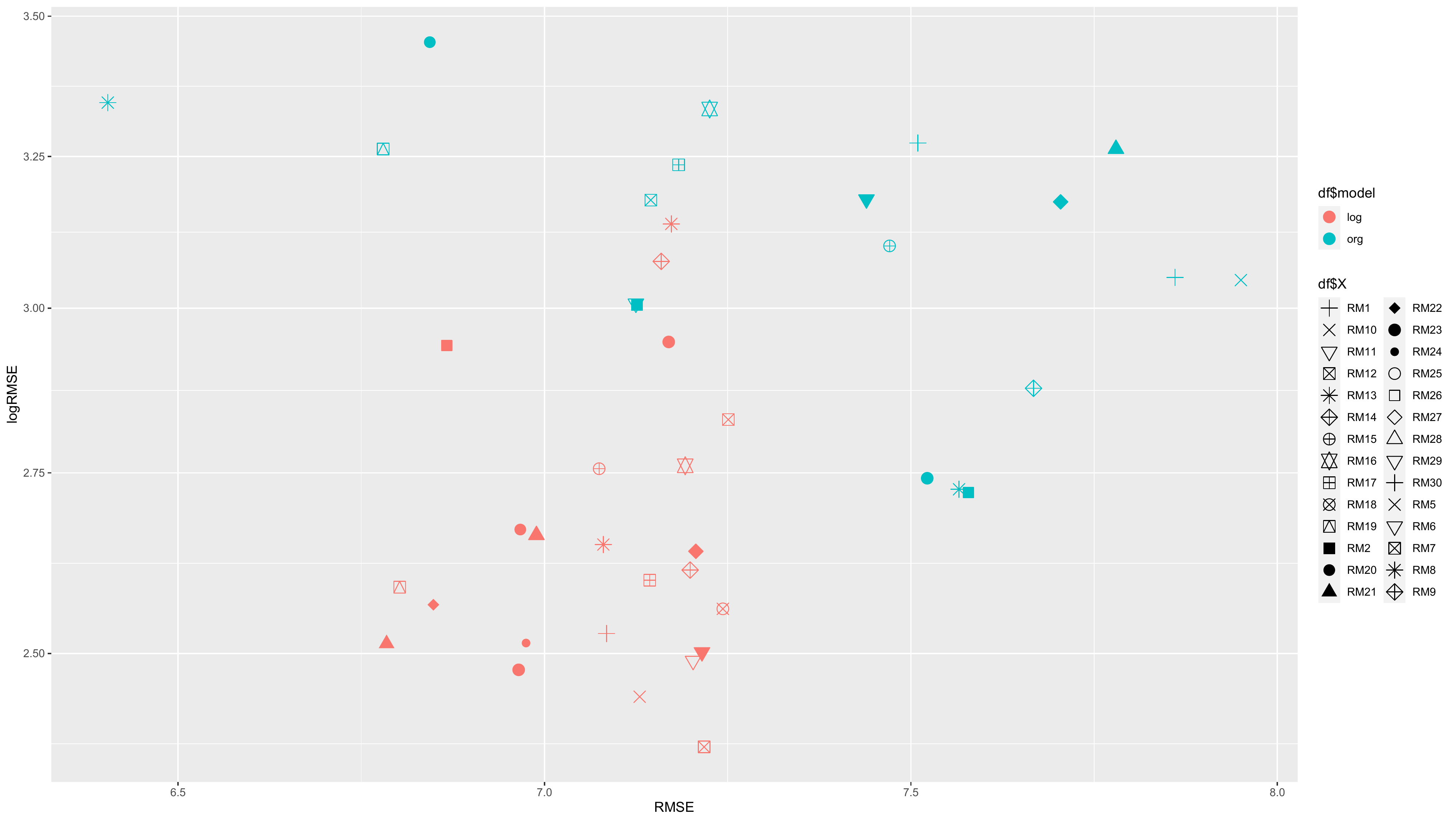}
    \caption{Overall prediction quality of different regression models (RMSE vs. log-RMSE), from both the unscaled (in blue) and the log-approach (in red) on the CMA-ES-CSA algorithm, with the budget of 1000 function evaluations and 2000-sample feature size. Minimizing both error values as a Pareto front, we see that log-based regression models perform significantly better than the unscaled ones.}
    \label{fig:regression-quality}
\end{figure*}

\subsection{Performance Regression Quality of Different Models}
\label{ssec:perf-pred}

Using the key elements described in Section~\ref{ssec:exp-setup}, we establish two separate regression (true and log-) approaches for each of the 30 regression models, as suggested in~\cite{JD20}. The intuition behind doing so lies in the fact that we want to make use of the information captured in the exponent of the actual performance value, which can be interpreted as a \emph{distance level} to the optimum (for two algorithms with actual target precisions of $10^{-2}$ and $10^{-8}$, we can say that the latter is 6 distance levels closer to the optimum than the former). In the remainder of the paper, we call these 2 approaches the unscaled and the log-approach, and we note that the unscaled one performs better in predicting target precisions ``far away from'' the optimum, while the log-one is more appropriate when targeting fine-grained cases which are already very close to the optimum.

Following common ML practices, we perform a 5-fold stratified \emph{leave-one-instance-out} cross-validation when training each model to reduce variability and obtain a higher model accuracy, thus carrying out the training on 4 out of 5 instances per each function, testing on the remaining one and combining the results over all folds.

In order to assess and compare the accuracy of different predictions made by the regression models, we compute the \emph{Root Mean Square Error} (RMSE) and its counterpart log-RMSE values per prediction, and aggregate them across problems and algorithms. The RMSE is a frequently used metric in the ML community and represents the standard deviation of prediction errors; put differently, it measures how spread out those errors are. In our fixed-budget regression, the prediction errors are the distance of the prediction to the true target value. Within this context, the RMSE is computed based on the actual, `un-logged' data (both for the unscaled and log-approach), whereas the log-RMSE makes use of the `logged' data.

Table~\ref{tab:rmse-predictions} conveniently shows the best achieved prediction accuracies (RMSE and log-RMSE) of the log-approach for different algorithms in the portfolio, for the 2000-sample feature size due to space constraints, for all different budgets, aggregated across problems. Different `Model' columns correspond to the regression model with the best quality in a specific scenario. Note that the RM-labeled models 1--6 refer to the Decision Tree regressors, 7--18 to the Random Forest regressors, and 19--30 to the Bagging DT regressors. This allows us to draw some first conclusions related to selecting the most efficient regressor, with the Decision Tree family being predominantly chosen in the low-budget setting when optimizing the RMSE, while the presence of the Bagging DT family is stronger when optimizing the log-RMSE in the same setting. Random Forest regressors, however, seem well-performing across the board, which is supported by the fact they are typically an all-round model of choice in related lines of research.  Most importantly, we remark that, even when aggregated across problems, the choice of the best regression model is highly dependent on the setting we work in, and varies between algorithms.

\begin{table*}[ht]
\caption{Best quality of the log-predictions (RMSE vs. log-RMSE) for each algorithm from the portfolio, for feature size of 2000 samples. Note that the labels in columns named `Model' correspond to the regression model achieving the said best prediction quality. Different models are the best-performing ones across the portfolio for different budgets of function evaluations.}
\label{tab:rmse-predictions}
\centering
    \resizebox{0.8\textwidth}{!}{

\begin{tabular}{@{}l|cccccccccccc@{}}
\toprule
             & \multicolumn{12}{c}{Feature size 2000}                                                                                                                                                                              \\ \midrule
             & \multicolumn{4}{c|}{Budget 250}                                              & \multicolumn{4}{c|}{Budget 500}                                              & \multicolumn{4}{c}{Budget 1000}                      \\ \midrule
Algorithm    & RMSE     & \multicolumn{1}{c|}{Model} & logRMSE & \multicolumn{1}{c|}{Model} & RMSE     & \multicolumn{1}{c|}{Model} & logRMSE & \multicolumn{1}{c|}{Model} & RMSE     & \multicolumn{1}{c|}{Model}   & logRMSE & Model \\ \midrule
BIPOP-CMA-ES & 10368.95 & \multicolumn{1}{c|}{RM27}  & 1.65    & \multicolumn{1}{c|}{RM13}  & 1489.86  & \multicolumn{1}{c|}{RM9}   & 1.83    & \multicolumn{1}{c|}{RM20}  & 73.18    & \multicolumn{1}{c|}{RM7}  & 1.94  & RM12   \\
BrentSTEPqi  & 58096.65 & \multicolumn{1}{c|}{RM8}   & 2.22    & \multicolumn{1}{c|}{RM30}  & 57828.22 & \multicolumn{1}{c|}{RM9}   & 2.76    & \multicolumn{1}{c|}{RM12}  & 57040.98 & \multicolumn{1}{c|}{RM9}  & 2.84  & RM11   \\
BrentSTEPrr  & 59475.56 & \multicolumn{1}{c|}{RM3}   & 2.15    & \multicolumn{1}{c|}{RM30}  & 58039.56 & \multicolumn{1}{c|}{RM16}  & 2.70    & \multicolumn{1}{c|}{RM30}  & 57993.93 & \multicolumn{1}{c|}{RM21} & 2.90  & RM12   \\
CMA-ES-CSA   & 10216.53 & \multicolumn{1}{c|}{RM7}   & 1.62    & \multicolumn{1}{c|}{RM13}  & 280.26   & \multicolumn{1}{c|}{RM13}  & 1.95    & \multicolumn{1}{c|}{RM30}  & 6.78     & \multicolumn{1}{c|}{RM21} & 2.38  & RM12   \\
fmincon      & 17.73    & \multicolumn{1}{c|}{RM5}   & 1.64    & \multicolumn{1}{c|}{RM18}  & 10.36    & \multicolumn{1}{c|}{RM3}   & 1.96    & \multicolumn{1}{c|}{RM18}  & 6.23     & \multicolumn{1}{c|}{RM20} & 1.64  & RM23   \\
fminunc      & 254.93   & \multicolumn{1}{c|}{RM21}  & 2.20    & \multicolumn{1}{c|}{RM11}  & 7.05     & \multicolumn{1}{c|}{RM7}   & 2.34    & \multicolumn{1}{c|}{RM29}  & 6.83     & \multicolumn{1}{c|}{RM2}  & 2.35  & RM29   \\
HCMA         & 1651.12  & \multicolumn{1}{c|}{RM13}  & 1.70    & \multicolumn{1}{c|}{RM23}  & 4.29     & \multicolumn{1}{c|}{RM23}  & 2.10    & \multicolumn{1}{c|}{RM18}  & 2.82     & \multicolumn{1}{c|}{RM14} & 2.77  & RM12   \\
HMLSL        & 16.83    & \multicolumn{1}{c|}{RM2}   & 1.62    & \multicolumn{1}{c|}{RM11}  & 10.45    & \multicolumn{1}{c|}{RM6}   & 1.95    & \multicolumn{1}{c|}{RM23}  & 4.01     & \multicolumn{1}{c|}{RM24} & 1.87  & RM12   \\
IPOP400D     & 3120.75  & \multicolumn{1}{c|}{RM14}  & 1.72    & \multicolumn{1}{c|}{RM18}  & 1204.75  & \multicolumn{1}{c|}{RM19}  & 1.71    & \multicolumn{1}{c|}{RM11}  & 33.66    & \multicolumn{1}{c|}{RM7}  & 1.99  & RM11   \\
MCS          & 2514.61  & \multicolumn{1}{c|}{RM2}   & 2.40    & \multicolumn{1}{c|}{RM24}  & 2233.27  & \multicolumn{1}{c|}{RM7}   & 2.74    & \multicolumn{1}{c|}{RM11}  & 1992.51  & \multicolumn{1}{c|}{RM24} & 2.87  & RM11   \\
MLSL         & 19.54    & \multicolumn{1}{c|}{RM7}   & 1.77    & \multicolumn{1}{c|}{RM17}  & 11.45    & \multicolumn{1}{c|}{RM6}   & 2.06    & \multicolumn{1}{c|}{RM23}  & 5.39     & \multicolumn{1}{c|}{RM6}  & 2.07  & RM22   \\
OQNLP        & 26.36    & \multicolumn{1}{c|}{RM13}  & 2.40    & \multicolumn{1}{c|}{RM30}  & 11.34    & \multicolumn{1}{c|}{RM20}  & 2.17    & \multicolumn{1}{c|}{RM17}  & 9.10     & \multicolumn{1}{c|}{RM20} & 2.29  & RM17   \\ \bottomrule
\end{tabular}
}
\end{table*}

As illustrated in the example in Figure~\ref{fig:regression-quality}, which is the use-case of CMA-ES-CSA algorithm for budget of 1000 evaluations and 2000-sample feature size, the overall regression quality both in terms of RMSE (on $x$-axis) and log-RMSE (on $y$-axis), regarded as a Pareto front, i.e. a two-objective min-min problem, is higher for the log-based approach than for the unscaled one.  However, it does not always have to be the case; depending on the algorithm, we can also observe situations in which the unscaled approach yields better results. Nevertheless, a remarkable diversity of regression model accuracies on different problems is easily noticed, and supports the idea of possibly having to resort to multiple regressors (each excelling in prediction of one sub-class of problems, for example) to achieve best results, which further supports the claim that ensemble-based models could be significantly more accurate than the standalone ones.

\section{ELA-based Algorithm Selection}
\label{sec:algo-selection}

After examining the regression accuracy, we proceed to evaluate the performance of two simple algorithm selectors, based on the predictions of the unscaled and the logarithmic approach, respectively, for each of the 30 regression models. The unscaled selector recommends the algorithm for which the unscaled-based model predicted the best performance, and, similarly, the log-selector bases its decision on the best prediction of the log-model. To quantify how well selectors perform per problem instance, we compare the precision of the algorithm chosen by the selector for the instance at hand to the precision of the actual best algorithm for that instance. We then indicate the overall quality of this selector by computing the RMSE and log-RMSE values (aggregated across all problem instances).

\begin{figure*}
    \centering 
    \includegraphics[trim=4cm 20cm 10cm 3cm, clip, width=0.7\linewidth, keepaspectratio]{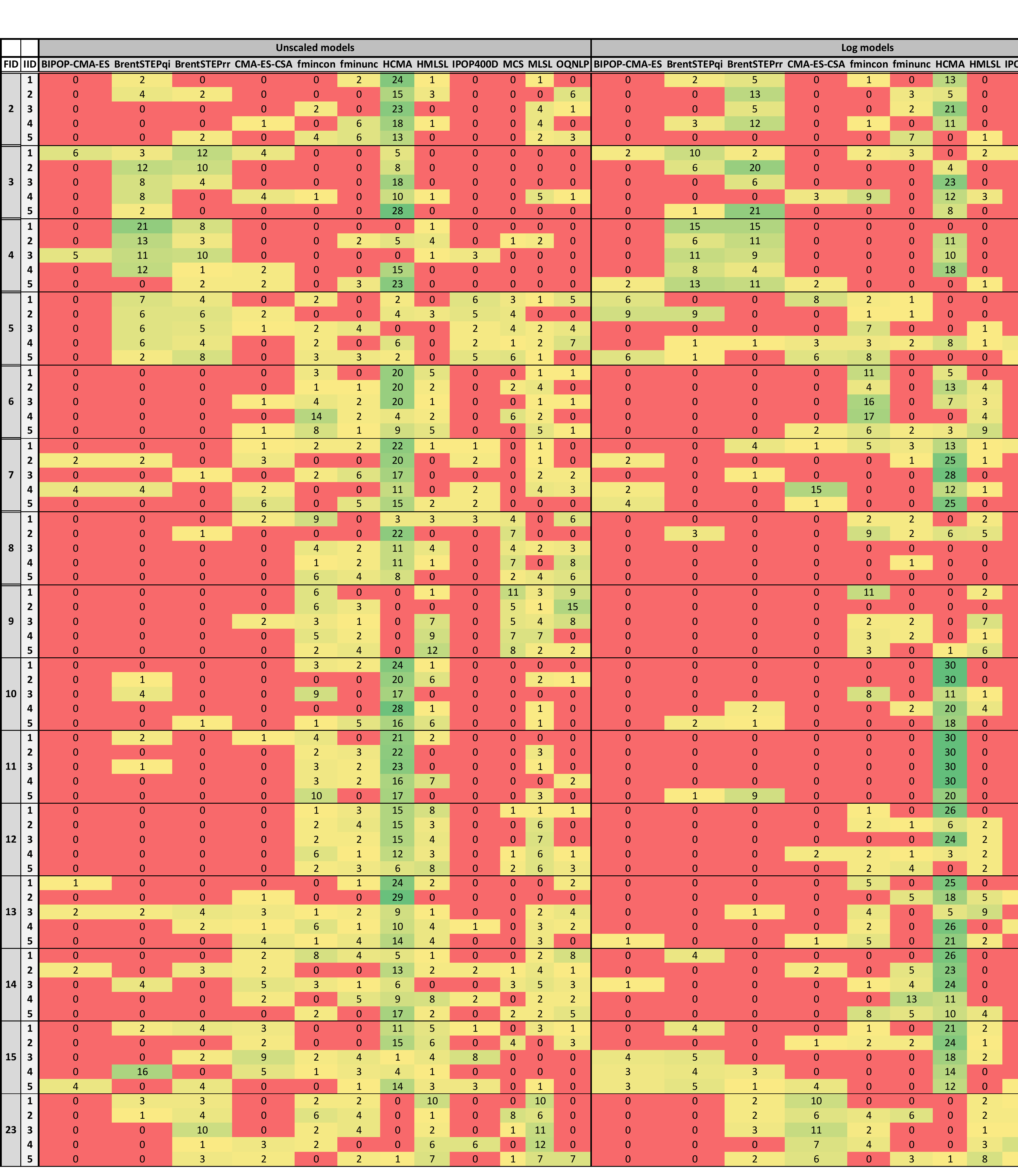}
    \caption{Heatmap of the selection frequency of each algorithm from the portfolio per problem instance, both for the unscaled (left) and log-approach (right), for the budget of 1000 function evaluations and the feature size of 250 samples, showing selected functions only for reasons of space.}
    \label{fig:heatmap}
\end{figure*}

Both in unscaled as well as in the log-approach, some algorithms will be selected more often than others per different problem instance, as seen in Figure~\ref{fig:heatmap}. For the budget of 1000 function evaluations and 2000-sample feature size, we observe that HCMA is chosen most consistently across different problem instances, while the choice of BIPOP-CMA-ES is very rare.  We also notice that on specific instances, there could be one or two particularly frequently selected (thus good) algorithms which are not selected for other problems in the benchmark set (e.g., CMA-ES-CSA for the function 16 in the unscaled approach, or fmincon for the function 21 in the log-approach).

\paragraph{\textbf{Combined Algorithm Selector}}

Following common practices in algorithm selection~\cite{bischl_aslib:_2016}, we compare the qualities of our two selectors (unscaled and log-based) with two different baselines -- one of them being the \emph{virtual best solver} (VBS, also called the oracle), which gives a lower bound as it always chooses the best-performing algorithm per each problem instance without any additional cost, so it reflects the best performance that could be achieved theoretically. The other baseline is the \emph{single best solver} (SBS), which stands for the overall (aggregated) performance of the best-performing algorithm from the portfolio. The SBS baselines vary depending on the budget and feature size, and we can also distinguish between the unscaled-approach SBS and the log-approach SBS.

\begin{figure*}
    \centering
    \includegraphics[width=\linewidth]{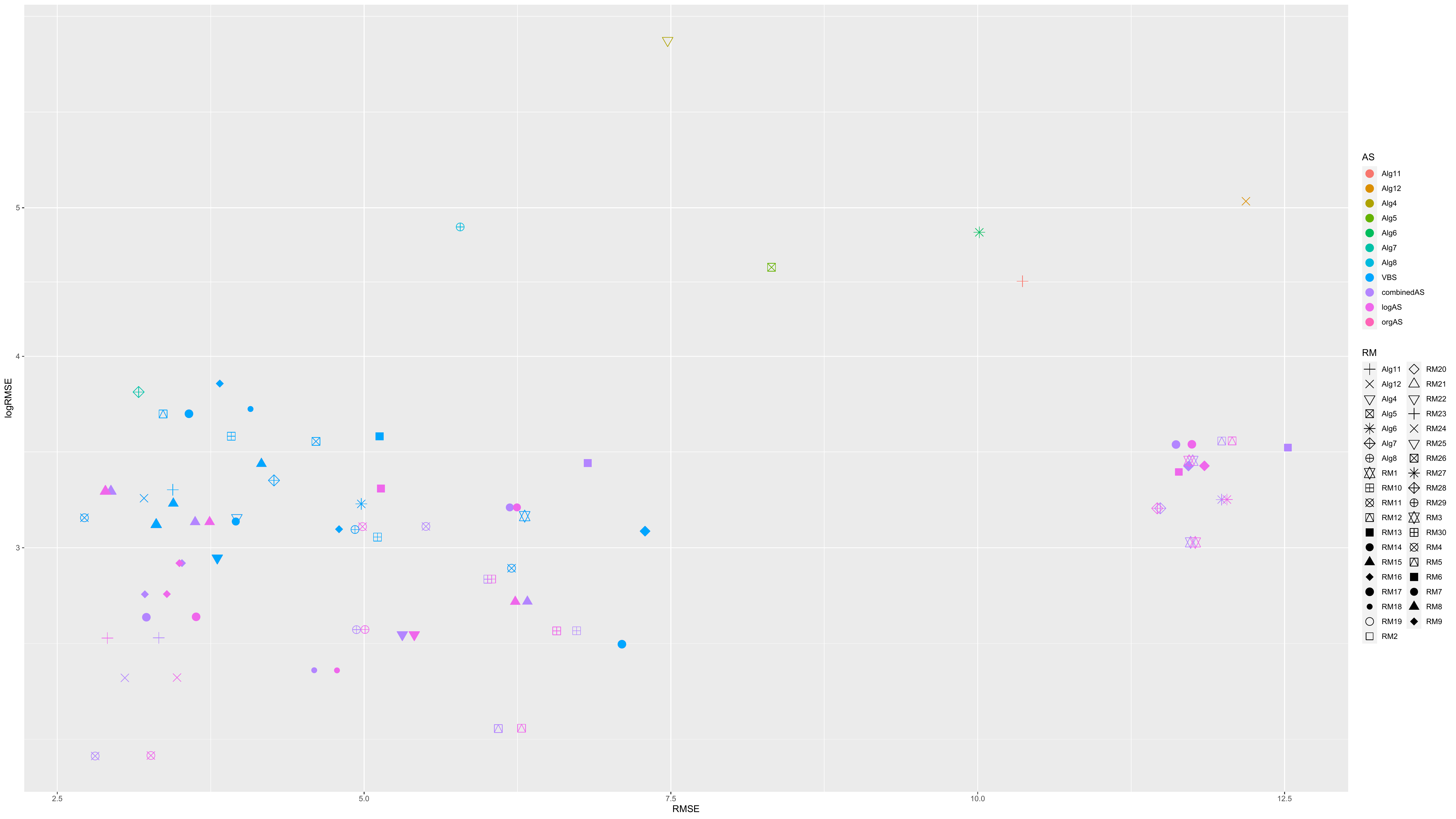}
    \caption{Quality of the algorithm selectors (RMSE vs. log-RMSE) for different regression approaches, including the combined selectors, the virtual best combined selectors (combinedVBS), and the selectors based on consistently using a certain algorithm across the board (`Algo' selectors in the legend, among which we can identify the SBS), for the budget of 1000 function evaluations and the 2000-sample feature size. Note that some of the selectors, notably the `Algo' ones, have been excluded from the figure as otherwise it disrupted the visibility of the best ones.}
    \label{fig:selectors}
\end{figure*}

We plot in Figure~\ref{fig:selectors} the quality of the selectors over all problem instances on RMSE and log-RMSE axes, again treating them as a Pareto front with the objective of minimizing both errors, and we see that different unscaled- and log-based selectors already outperform the majority of single algorithms from the portfolio on both RMSE and log-RMSE.  We also want to incorporate the observation reported in Section~\ref{ssec:perf-pred} that the unscaled approach is better in predicting higher target precision, while the log-approach has better accuracy when targeting smaller performance values. 

We hence establish a combined regression approach for all the models in order to benefit from their two complementing strengths. Here again we can define a combined VBS, which is the one that, for each model and problem instance, chooses the better of the two recommended algorithms by the unscaled- and the logarithmic approach, respectively, using the following rule: if the target precision of an algorithm, as predicted by the log-model, is smaller than a certain threshold, we use the log-approach, whereas we use the recommendation of the unscaled approach otherwise.  The threshold value chosen here is 0.9 in order to ensure selecting the log-approach recommendation for fine-grained precisions and vice versa. Note that that a complete sensitivity analysis with respect to the threshold value was described in~\cite{JankovicD20}, where ad-hoc threshold optimization was performed for each regression model in order to minimize RMSE and log-RMSE.  Due to space constraints, the sensitivity analysis was out of scope of this work.  We apply this strategy to all 30 regression models, and the results for the budget of 1000 evaluations and the 2000-sample feature size are seen in Figure~\ref{fig:selectors}. The combined approach clearly outperforms any of the simple regression approaches, the single algorithms and even the combined VBS selectors. This finding highlights the potential of the combined selector, which boosts the quality of its standalone components even in the case when they might not be the optimal regression models for a specific algorithm portfolio and s problem set.

\section{Sensitivity Analysis}
\label{sec:sens-analysis}

Lastly, we highlight the differences in regression quality obtained by using different budgets and feature sizes. We have purposefully randomly selected one regression model (\emph{RandomForest crit.mse minsplit.4 nest.9}) and one algorithm (HCMA) to point out the diversity of regression on different problem instances. The data in Table~\ref{tab:sens-analysis} corresponds to the log-approach of said selector, using 2000-sample feature size, in order to be consistent in terms of the showcased use-case throughout the paper. We immediately observe extremely large RMSE values for the budget of 250 function evaluations, which drastically decrease for the budgets of 500 and 1000. One possible interpretation of this finding could be that in raw performance data, whereas the budget of 250 evaluations was not big enough to allow for getting closer to the optimum in case of certain functions, increasing the budget seem to facilitate getting a better estimate of the optimal solution, thus making a more straightforward way for the regression to perform better. Finally, this again stresses the possible importance of personalizing the regression models to the actual optimization problems, as the results can vary drastically between problem instances.

\begin{table*}[h!]
\caption{Sensitivity analysis of regression quality for the log-approach of the $RandomForest_crit.mse_minsplit.4_nest.9$ regression model for the HCMA algorithm for the feature size of 2000 samples, with respect to the different algorithm budgets. Note that, for reasons of space, we showcase only values of the first instance of the 24 BBOB functions.}
\centering

\resizebox*{\textwidth}{!}{
\begin{tabular}{@{}lcc|cccccccccccccccccccccccc@{}}
\toprule
\multicolumn{3}{c}{FID}                                                                                               & 1    & 2     & 3     & 4    & 5     & 6      & 7     & 8        & 9    & 10          & 11      & 12          & 13    & 14    & 15     & 16     & 17   & 18    & 19   & 20   & 21   & 22   & 23   & 24     \\ \midrule
\multicolumn{3}{c|}{IID}                                                                                              & 1    & 1     & 1     & 1    & 1     & 1      & 1     & 1        & 1    & 1           & 1       & 1           & 1     & 1     & 1      & 1      & 1    & 1     & 1    & 1    & 1    & 1    & 1    & 1      \\ \midrule
\multicolumn{1}{l|}{\multirow{6}{*}{Feature size 2000}} & \multicolumn{1}{c|}{\multirow{2}{*}{Budget 250}}  & RMSE    & 0.12 & 0.00  & 4.01  & 0.66 & 1.00  & 229.24 & 39.73 & 44127.00 & 0.01 & 21357821.80 & 3243.65 & 20853295.16 & 53.95 & 1.12  & 656.09 & 341.45 & 7.21 & 38.76 & 0.09 & 6.81 & 0.49 & 0.72 & 3.89 & 247.95 \\
\multicolumn{1}{l|}{}                                   & \multicolumn{1}{c|}{}                             & logRMSE & 0.03 & 15.35 & 0.01  & 0.00 & 14.80 & 0.40   & 0.21  & 1.12     & 0.00 & 1.94        & 0.04    & 0.65        & 0.11  & 1.19  & 0.19   & 0.58   & 0.40 & 0.18  & 0.12 & 0.15 & 0.03 & 0.06 & 0.09 & 0.06   \\ \cmidrule(l){2-27} 
\multicolumn{1}{l|}{}                                   & \multicolumn{1}{c|}{\multirow{2}{*}{Budget 500}}  & RMSE    & 0.06 & 0.00  & 0.16  & 1.77 & 1.00  & 4.83   & 0.52  & 10.35    & 0.13 & 0.00        & 0.00    & 0.00        & 0.00  & 0.01  & 251.70 & 72.77  & 0.11 & 4.18  & 0.03 & 0.84 & 0.70 & 0.26 & 0.55 & 5.75   \\
\multicolumn{1}{l|}{}                                   & \multicolumn{1}{c|}{}                             & logRMSE & 0.01 & 23.83 & 7.65  & 0.67 & 21.68 & 1.98   & 4.35  & 0.28     & 0.10 & 1.09        & 1.50    & 0.02        & 0.67  & 13.13 & 6.62   & 0.39   & 1.09 & 2.49  & 0.06 & 0.09 & 0.04 & 0.02 & 0.02 & 0.01   \\ \cmidrule(l){2-27} 
\multicolumn{1}{l|}{}                                   & \multicolumn{1}{c|}{\multirow{2}{*}{Budget 1000}} & RMSE    & 0.01 & 0.00  & 2.58  & 0.66 & 1.00  & 0.00   & 0.41  & 0.00     & 0.00 & 0.00        & 0.00    & 0.00        & 0.00  & 0.00  & 0.99   & 0.10   & 0.00 & 0.18  & 0.14 & 2.86 & 0.10 & 3.72 & 0.23 & 1.65   \\
\multicolumn{1}{l|}{}                                   & \multicolumn{1}{c|}{}                             & logRMSE & 0.00 & 0.02  & 80.68 & 2.80 & 11.15 & 1.28   & 16.98 & 0.01     & 3.85 & 0.04        & 0.53    & 0.26        & 0.15  & 1.01  & 33.19  & 0.02   & 2.96 & 2.86  & 0.15 & 1.48 & 0.02 & 3.48 & 0.01 & 0.00   \\ \bottomrule
\end{tabular}
}
\label{tab:sens-analysis}
\end{table*}

\section{Discussion, Conclusions and Future Work}
\label{sec:conclusion}

While most landscape-aware algorithm selection, design, and configuration studies in the context of numerical black-box optimization do not pay great attention to the configuration of the ML model, we have demonstrated in this work that both the choice of regression technique and its parametrization can have significant impact on the performance of the trained models. Using a classical experimental design from the context of automated algorithm selection~\cite{KerschkeT19,JankovicD20}, we have analyzed 30 different regression models, applied them to both the log- and to the unscaled performance data, trained an automated algorithm selector, and analyzed its performance through stratified 5-fold cross validation.  We have seen that the regression quality of the different models can vary by several orders of magnitude. This reinforces and justifies the need for meticulously choosing the ML model and its hyper-parameter configuration, as we have seen that picking any single model and tuning it might not at all provide good results (i.e., the discarded ML models from the preliminary step of this work). Differences in the regression models' quality also lead to very diverse performance portfolios in the algorithm selection task, although the impact there is somewhat less severe, since wrong predictions can still result in a lucky choice of algorithms. Additionally, when considering ensembles, hyper-parameter tuning can be expected to further improve upon performance gains, which will depend on the used data set and learning scenario. 

Our study suggests that the selection of the ML techniques should be performed with care. It also suggests that the quality of different regression techniques can vary between different types of problems, so that we cannot give a ``one size fits all'' recommendation for which regression models or which parametrization to favor. As a rule of thumb, different BaggingDT and RandomForest instances provide better results in terms of log-RMSE, for both feature extraction sample sizes and all three budgets of function evaluations. For the RMSE performance criterion, in contrast, DecisionTrees provides best results for some of the algorithms. The important question of choosing the absolute best performing ML model for a certain problem set and algorithm portfolio remains open, but this preliminary work stresses the significance of the efforts that should go towards developing more advanced mechanisms to select the most appropriate one and its hyper-parameter configuration.

We note that the computational costs of training different regression models is rather negligible for the data sizes commonly studied in the landscape-aware black-box algorithm selection and configuration context, which means a validation step can be added before deciding which model to choose and apply to the real use-cases (i.e., in the \emph{test} phase). In general, we believe that the current common practice of studying stratified 5-fold cross validation on the BBOB functions is too limited to give an accurate impression about the potential of carefully choosing the ML model for the automated algorithm selection and configuration in practice. We therefore plan to massively extend our study by adding to our data sets performance data from the Nevergrad platform~\cite{nevergrad}, which offers benchmark data for very broad ranges of optimization problems on its frequently updated dashboard.

One problem to overcome in cross-validation across different benchmarking platforms is the fact that one needs to ensure that the data corresponds to the same instances of the algorithms -- a ``CMA-ES'' in one platform may be much different than a CMA-ES implementation in another. We therefore believe that a common algorithm repository, interfaced with the various benchmarking suites, would be a useful step towards a better re-usability of results and better training sets for the automated design, selection, and configuration of black-box optimization techniques.

\paragraph{\textbf{Acknowledgments}}
 This work has been supported by the Paris Ile-de-France region, and the Slovenian Research Agency (research core funding No. P2-0098, project No. Z2-1867, and grant number PR-10465).
\bibliographystyle{ACM-Reference-Format}
\bibliography{references}

\end{document}